# Dimension Reduction by Mutual Information DISCRIMINANT ANALYSIS

Ali Shadvar[1]

[1]Department of Biomedical Engineering, Science and Research Branch, Islamic Azad University, Langrud, Iran
Shadvar@iaul.ac.ir

## ABSTRACT

*In the past few decades, researchers have proposed many discriminant analysis (DA) algorithms for the study of high-dimensional data in a variety of problems. Most DA algorithms for feature extraction are based on transformations that simultaneously maximize the between-class scatter and minimize the within-class scatter matrices. This paper presents a novel DA algorithm for feature extraction using mutual information (MI). However, it is not always easy to obtain an accurate estimation for high-dimensional MI. In this paper, we propose an efficient method for feature extraction that is based on one-dimensional MI estimations. We will refer to this algorithm as mutual information discriminant analysis (MIDA). The performance of this proposed method was evaluated using UCI databases. The results indicate that MIDA provides robust performance over different data sets with different characteristics and that MIDA always performs better than, or at least comparable to, the best performing algorithms.*

## KEYWORDS

*Classification, Discriminant analysis, Dimension reduction, Feature extraction, Mutual information*

## 1. INTRODUCTION

Dimensionality reduction of the raw input variable space is an essential preprocessing step in the classification process. There are two main reasons to keep the dimensionality of the input features as small as possible: computational cost and classification accuracy. It has been observed that added irrelevant features may actually degrade the performance of classifiers if the number of training samples is small relative to the number of features [1-2].

Reduction of the number of input variables can be achieved by selecting relevant features (i.e., feature selection), [3-4] or by extracting new features that contain maximal information about the class label from the set of original features (i.e., feature extraction) [5-6]. To keep some of the original features, it may be more suitable to perform feature selection. However, when the number of irrelevant features is orders of magnitude larger than the number of relevant features, feature selection requires a large set of training data to obtain reliable transformations. Additionally, because switching from one feature to another is a discrete operation, feature selection is not a smooth process. Another reason that motivates using feature extraction over selection is feature extraction's power to distribute relevant information amongst different original features. This advantage results in greater information compaction [7].

The function that describes feature extraction is $z = f(x)$, $z \in R^N$. This function can be either linear or nonlinear. The classifier determines whether a linear or nonlinear extraction method ought to be used. For this reason, deciding whether to apply a nonlinear feature extraction method before a linear classifier, or a linear classifier followed by nonlinear classifier, is common. In the first case, the input data are projected by a nonlinear feature extractor onto a set of variables. The nonlinear patterns are unfolded, and the separation of the classes is made possible by a linear classifier. In the second case, the classifier has the responsibility of finding the nonlinear separation boundaries [8]. In [9], it has been shown that a proper linear transformation on input data improves the performance of the simple k-nearest-neighbors (KNN) classifier. For this reason, we will consider linear feature extraction in this paper.

Currently, many researchers have presented a variety of methods for linear feature extraction. PCA is one of the most well known methods. Finding an orthogonal set of projection vectors for extracting features is the ultimate goal of PCA. PCA extracts features by maximizing the variance of data. Though this approach is an adequate method for reducing dimensionality, because of its unsupervised nature, it is not suitable for classification-based feature extraction tasks [10].
Linear discrimination analysis (LDA) is also a well known and popular linear-dimension-reduction algorithm for supervised feature extraction [11]. LDA computes a linear transformation by maximizing the ratio of between-class distance to within-class distance, thereby achieving maximal discrimination. In LDA, a transformation matrix from an n-dimensional feature space to a d-dimensional space is determined such that the Fisher criterion of between-class scatter over within-class scatter is maximized.

However, the traditional LDA method is based on the restrictive assumption that the data are homoscedastic, *i.e.,* the classes of the data have equal covariance matrices. In particular, it is assumed that the probability density functions of all classes are Gaussian with identical covariance matrices, but with different means [12]. Moreover, traditional LDA cannot solve the problem posed by nonlinearly separable classes. Hence, LDA's performance is unsatisfactory for many classification problems that have nonlinear decision boundaries. To solve this problem, a nonlinear extension of LDA has been proposed [13-14].

Moreover, LDA-based algorithms generally suffer from the small sample size (SSS) problem that occurs when the number of training samples is less than the dimension of feature vectors [15]. A traditional solution to this problem is to apply PCA in conjunction with LDA [16]. Recently, more effective solutions have been proposed to solve the SSS [10].

Another problem that is common to most DA methods is that these methods can only extract *C*-1 features from the original feature space, where *C* is the number of classes. Recently, a method based on DA was proposed for describing a large number of data distributions and to solve the limitation posed by DA methods on the number of features that can be extracted. This method is known as subclass discriminant analysis (SDA) [17].

One of the most effective approaches for optimal feature extraction is based on MI. MI measures the mutual dependence of two or more variables. In this context, the feature extraction process is creating a feature set from the data that jointly have the largest dependency on the target class and that has minimal redundancy. However, it is almost impossible to obtain an accurate estimation for high-dimensional MI. In [7, 18], a method known as MRMI was proposed for learning linear discriminative feature transforms with an approximation of the MI between transformed features and class labels as a criterion. The approximation is inspired by the quadratic Renyi entropy, which provides a nonparametric estimate of the MI. However, there is no general guarantee that maximizing the approximation of MI using Renyi's definition is equivalent to maximizing MI defined by Shannon. Moreover, the MRMI algorithm is subject to the curse of dimensionality. In [8], a method of extracting features based on one-dimensional MI has been presented. This method is called MMI. In this method, the first feature is extracted in a manner that maximizes the MI between the extracted feature and the class of the data. The other features must be extracted such that they are orthogonal and maximize the MI between the extracted features and the class label. However, in general the orthogonality of a newly extracted feature relative to previous features cannot guarantee independence; as such, this method still cannot eliminate redundancy. To overcome the difficulties of MI estimation for feature extraction, Parzen window modeling has been employed to estimate the probability density function [19]. However, the Parzen model may suffer from the "curse of dimensionality," which refers to the over-fitting of high-dimensional training data [8].

The purpose of this paper is to introduce an efficient method for extraction of features with maximal dependence on the target class and minimal redundancy between extracted features. This method uses one-dimensional MI estimation and Fisher-Rao's criterion to overcome the practical obstacles mentioned above. The proposed method is then evaluated against six databases. The results were compared with those obtained from the PCA-, LDA-, SDA- [17] and MI-based feature extraction methods (MRMI-SIG) proposed in [18]. The results indicate that MIDA provides robust performance over different data sets with different characteristics and that MIDA always performs better than, or at least comparable to, the best algorithms.

The rest of the paper is divided as follows. In Section II, a summary of information theory concepts is provided. In Section III, we describe our algorithm for feature extraction. In Section IV, based on experiments, we compare the practical results of our method with other methods. In Section V, we conclude.

## 2. BACKGROUND ON INFORMATION THEORY

### 2.1. Mutual information and feature extraction

MI is a nonparametric measure of relevance between two variables. Shannon's information theory provides a suitable formalism for quantifying these concepts [20]. Assume that a random variable $X$ represents a continuously valued random feature vector and that a discrete-valued random variable $C$ represents the class labels. In accordance with Shannon's information theory, the uncertainty of class label $C$ can be measured by entropy $H(C)$ as

$$H(C) = -\sum_{c \in C} p(c) \log p(c) \qquad (1)$$

where p($c$) represents the probability of the discrete random variable $C$. The uncertainty of $C$ given a feature vector $X$ is measured by the conditional entropy as s

$$H(C|X) = -\int_X p(x) \left( \sum_{c \in C} p(c|x) \log p(c|x) \right) dx \qquad (2)$$

where $p(c|x)$ is the conditional probability for variable $C$ given $X$.
In general, the conditional entropy is less than or equal to the initial entropy. The conditional entropy is equal if and only if variables $C$ and $X$ are independent. By definition, the amount that the class uncertainty is decreased by is the MI. As such, $I(X;C) = H(C) - H(C|X)$. After applying the identities $p(c,x) = p(c|x) p(x)$ and $p(c) = \int_x p(c,x) dx$, $I$ can be expressed as

$$I(X;C) = \sum_{c \in C} \int_x p(c,x) \log \frac{p(c,x)}{p(c) p(x)} dx \qquad (3)$$

If the MI between two random variables is large, then the two variables are closely related. The MI is zero if and only if the two random variables are strictly independent [21].

In classification problems, suitable features are those that have a higher quantity of MI with respect to classes. There are two bounds on Bayes error that justify the use of MI for feature extractions. The first bound is Hellman and Raviv's upper bound: $p_e \leq (H(C) - I(X;C))/2$. The second bound is Fano's lower bound: $p_e \geq (H(C) - I(X;C) - 1)/\log(N_c)$. As the MI grows, the bounds decrease and reduce Bayes error. So using MI is a reasonable criterion for feature extraction. On the other hand, according to the inequality of data processing for any deterministic transformation $T(\cdot)$, we hold

$$I(T(x);C) \leq I(X;C) \qquad (4)$$

This equality is only held when the transformation process is invertible [22], so that no improvement will occur in the MI existing between the data and classes. For this reason, our objective in this paper is to propose a heuristic method for feature extraction that is based on a minimal-redundancy-maximal-relevance framework that maximizes the information in a reduced space.

## 2.2. Discriminant Analysis

Most of the previously defined DA methods are based on Fisher-Rao's criterion [23], which is given by

$$V = \arg\max \frac{|V^T A V|}{|V^T B V|} \qquad (5)$$

Matrices $A$ and $B$ are assumed to be symmetric and positive-definite so that they define a metric. LDA, respectively uses the between- and within-class scatter matrices $A = S_B$ and $B = S_W$ in (1).

$$S_B = \sum_{i=1}^{C} (\mu_i - \mu)(\mu_i - \mu)^T \qquad (6)$$

$$S_W = \frac{1}{n} \sum_{i=1}^{C} \sum_{j=1}^{n_i} (x_{ij} - \mu_i)(x_{ij} - \mu_i)^T \qquad (7)$$

$C$ is the number of classes, $\mu_i$ the sample mean of class $i$, $\mu$ the global mean, $x_{ij}$ is the $j$th sample of class $i$, and $n_i$ is the number of samples in that class. The objective is to find a linear transformation $V$ that maximizes the between-class scatter matrix $S_B$ and minimizes the within-class scatter matrix $S_W$ (Fisher's criterion). Other DAs are formed by redefining either the *A* or *B* matrices. As an example NDA, the following nonparametric version,
where $\alpha_{ij}^l$ is the parameter of the between-class scatter matrix for *A*, is used:

$$S_B = \frac{1}{n} \sum_{i=1}^{C} \sum_{j=1}^{n_i} \sum_{\substack{l=1 \\ l \neq i}}^{C} \alpha_{ij}^l (x_{ij} - M_{ij}^l)(x_{ij} - M_{ij}^l)^T, \qquad (8)$$

This nonparametric version avoids the outcomes that are affected by samples placed away from the boundary. $M_{ij}^l$ is the mean sample ($x_{ij}$) that belongs to class $l (l \neq i)$. In another DA algorithm known as aPAC, a new weighted scatter matrix is defined as $\sum_{i=1}^{C-1} \sum_{j=i+1}^{C} w(d_{ij}) S_{ij}$, where $S_{ij} = (\mu_i - \mu)(\mu_i - \mu)^T$, where $d_{ij}$ is the Mahalanobis distance between classes *i* and *j*, and $w(\cdot)$ is a weight function that provides the equality of each class contribution to the classification accuracy. PDA is another DA algorithm that is formed by redefining the within-scatter matrix as $B = S_W + \Omega$, where $\Omega$ is a penalizing matrix that penalizes noisy eigenvectors.

A common problem with most DA methods results from a deficiency in the rank of the between-scatter matrix. For example, the rank of $S_B$ in the LDA, PDA, aPAC, and most other DA methods is smaller than $C - 1$ ($rank(S_B) \leq C - 1$). These methods are able to extract only $C - 1$ features from the original feature space. These $C - 1$ features will be sufficient if the classes are separated linearly. However, for realistic data sets, most data cannot be separated linearly. Therefore, this limitation prevents these methods from extracting the optimum set of features [24]. To overcome this problem, in [25], heteroscedastic LDA is introduced as a new DA that redefines the *A* matrix such that it considers both the differences between class means and their covariances, as shown in equation 9.

$$S_C = \sum_{i=1}^{C-1} \sum_{j=i+1}^{C} \left[ \Sigma_{ij}^{-1/2} (\mu_i - \mu)(\mu_i - \mu)^T \Sigma_{ij}^{-1/2} + 4(\log \Sigma_{ij} - \frac{1}{2} \log \Sigma_i - \frac{1}{2} \log \Sigma_j) \right] \qquad (9)$$

where $\Sigma_i$ is the covariance matrix of the samples in class *i*, and $\Sigma_{ij}$ is the average between $\Sigma_i$ and $\Sigma_j$. In [17], another DA, subclass discriminant analysis (SDA), is proposed to overcome this problem. In the SDA approach, to increase the rank of $S_B$, the authors define a new *A* matrix by dividing each class into subclasses:

$$\Sigma_B = \sum_{i=1}^{C-1} \sum_{j=1}^{H_i} \sum_{k=i+1}^{C} \sum_{l=1}^{H_k} p_{ij} p_{kl} (\mu_{ij} - \mu_{kl})(\mu_{ij} - \mu_{kl})^T \qquad (10)$$

where $p_{ij}$ and $\mu_{ij}$ are the prior and mean of the *j*th subclass in class *i*, and $H_i$ is the number of subclass divisions in class i.

## 3. MUTUAL INFORMATION DISCRIMINANT ANALYSIS

To obtain an optimal extraction of features from the set of original features, we need to create a new set of features that has the largest dependency on the target class. Let us denote the original feature set as *X*, which is a sample of the continuously valued random vector. *C* is a discrete-valued random variable that represents the class labels. The problem is to find a linear mapping $W$ such that the transformed features

$$Y = W^T X \qquad (11)$$

Maximize the MI between the transformed features $Y$ and the class labels $C$, $I(W^T X; C)$. That is, we seek

$$W_{opt} = \arg\max_W I(W^T X; C) \qquad (12)$$

$$I(Y;C) = \sum_{c \in C} \int \ldots \int p(y_1 \ldots y_m) \log \frac{p(y_1 \ldots y_m, c)}{p(y_1 \ldots y_m) p(c)} \times dy_1 \ldots dy \qquad (13)$$

The requirement of knowing the underlying probability density functions (PDFs) of the data and the integration of these PDFs always makes it difficult to accurately estimate high-dimensional MI [26]. The above-mentioned solution is not practical because of its large computational requirements for complex problems.

To overcome the above-mentioned practical obstacles, in this paper we have used Fisher-Rao's criterion and one-dimensional MI estimation to estimate the MI for low-dimensional data spaces. We used a popular histogram method [27] to obtain the estimation. Histogram estimators can deliver satisfactory results for low-dimensional data spaces.

Most DA algorithms try to separate class means as well as possible, not taking into consideration discriminatory information. Moreover, because most DA algorithms only make use of second-order statistical information, the covariance is optimal for data that have a unimodal Gaussian density with well-separated means for each class. Furthermore, the maximum rank of $S_B$ is $C-1$. Thus, these methods cannot produce more than $C-1$ features. To overcome this problem, in this section, we define a new information-theory criterion based on Fisher-Rao's criterion and MI.

Consider a data set $X$ that is represented by $M$ samples $\{x_1, x_2, \ldots, x_M\}$. Each sample is represented by $N$ features $\{f_1, f_2, \ldots, f_N\}$ that take on values in $N$-dimensional space. Each sample also belongs to one of $C$ classes. The new between-class scatter matrix $S_B$ and the within-class scatter matrix $S_w$, based on the one-dimensional MI, can be defined as follows:

$$S_B = \sum_{i=1}^{N} \sum_{j=1}^{N} a_{ij} \qquad (14)$$

$$S_w = \sum_{i=1}^{N} \sum_{j=1}^{N} b_{ij} \qquad (15)$$

where $a_{ij}$ and $b_{ij}$ can be defined as follows

$$a_{ij} = \begin{cases} I(f_i; C) & i = j \\ 0 & i \neq j \end{cases} \qquad (16)$$

$$b_{ij} = \begin{cases} 0 & i = j \\ ct + I(f_i; f_j) & i \neq j \end{cases} \qquad (17)$$

International Journal of Artificial Intelligence & Applications (IJAIA), Vol.3, No.3, May 2012where $I(f_i;C)$ is the MI between the i-th feature, and the class label and *ct* are integer constants. When $S_B$ is maximized, the features with largest dependency on the target class are extracted. When $S_w$ is minimized, the features with minimum redundancy are extracted. *ct* is a parameter that determines how the extracted features are dependent relative to the class label, as well as how independent the extracted features are from other features. Then, *ct* determines how much $S_B$ should be maximized and how much $S_w$ should be minimized. These optimal values will be calculated as follows:

1) Initialization:
   Set *X* to the initial feature set;
   Set *S* to the empty set;
   Set *ct* to zero;
   Set t to the desired number of features that will be extracted
2) For ct=0 to L, do the following:
   Calculate $S_B$ using (14) and $S_W$ using (15)

$$W = \arg\max_V \frac{|V^T S_B V|}{|V^T S_W V|}$$

   Project the data

$$Y = W^T X$$

   Calculate

$$K = \sum_{i=1}^{t} \left\langle I(y_i;C) - \frac{1}{i-1}\sum_{j=1}^{i-1} I(y_i;y_j) \right\rangle$$

   End for
3) Finding the optimal value for *ct*

$$ct_{opt} = \arg\max_{ct} K$$

By solving optimization problem (5) using (14) and (15), the projection vector set consists of the eigenvectors that correspond to the nonzero eigenvalues of $B^{-1}A$:

$$AV = BV\Lambda \qquad (18)$$

The transformation matrix $W$ must be created from the largest eigenvectors $V$:

$$W = [v_1, v_2, v_3, \ldots v_t] \qquad (19)$$

The optimal feature set is obtained by projecting the original feature set onto the projection matrix:

$$Y = W^T X \qquad (20)$$

29

where $X$ is the original feature set, and $Y$ is the optimal feature set.

By solving the optimization problem, new features are extracted from the original features. These new features jointly have the largest dependency on the target class and minimal redundancy. The proposed MI-based feature extraction can be summarized by the following procedure:

1) Initialization:
    Set $X$ to the initial feature set;
    Set $S$ to the empty set;
    Set $ct$ to $ct_o$;
    Set $t$ to the desired number of features that will extracted
2) Determine the weighting matrix
    Calculate $S_B$ using (14) and $S_W$ using (15)

$$W_{opt} = \arg\max_V \frac{|V^T S_B V|}{|V^T S_W V|}$$

3) Extract the feature
$$Y = W_{opt}^T X$$
$$S \leftarrow \{Y\}$$
4) Output the set $S$ that contains the extracted features.

Table 1. Description of the data sets used in the comparison.

| Data set | Features | Classes | Samples |
| --- | --- | --- | --- |
| Letter | 16 | 26 | 20000 |
| Libras movement | 90 | 15 | 360 |
| Wall-following | 24 | 4 | 5456 |
| Madelon | 500 | 2 | 2600 |
| Hill-valley with noise | 100 | 2 | 1212 |
| Hill-valley without noise | 100 | 2 | 1212 |

## 4. RESULTS

In this section, we investigate the performance of the proposed method using several UCI data sets (the UCI machine learning repository contains many real-world data sets that have been used by a variety of investigators) [28] and compare the obtained results with other well-known feature extraction methods: PCA, LDA, SDA and the MI-based feature extraction method proposed in [18]. The MI-based method is also known as MRMI-SIG.

A support vector machine (SVM) [29] with a Gaussian radial basis function as a kernel and a KNN classifier [30] has been applied to evaluate the classification performance. A SVM classifier is usually picked because it is less sensitive to the curse of dimensionality than other classifiers. As such, the quantity of information that this data will carry about classes will be in high correlation with its performance. A tenfold cross-validation procedure on the training data is used to determine the cost and width of the SVM kernel. Instead, a KNN classifier with all of its simplicity performs so well in these types of experiments that it is often used to compare a variety of methods. It is because of its good performance that we applied KNN with K=1 in this paper. To obtain more reliable results, divide by the absolute maximum of the training set to normalize input values of the data for all classifiers.

To increase the significance of our result statistics obtained from using data sets with a limited number of samples, and to obtain the classification rates, the average values over 10-fold cross-validation have been applied. To assess the classification accuracy for every 10-fold partition, nine were used as a training set and one as test set. First, the algorithm for our feature extraction was run on our training sets. Then, the classifier was trained and tested. At the end, the average classification results were reported as the error. To evaluate the performance of our method presented in this paper, six data sets have been used. Table I shows brief information about the data sets used in this paper. In the following section, we will give a short description of each data set and the results obtained by examining those methods using KNN and SVM classifiers. Then, we compare the results of our algorithm with the results from the other methods.

The first data set used in our paper was the Letter data set. The objective is to identify the 26 capital letters of the English alphabet. This data set is consists of 20000 samples. Each sample comprises 16 attributes that were scaled to fit into the integer range from 0 through 15. Here, our classifiers showed different results. In the first component, KNN indicated that SDA was better by 22.2%. However, our method had adequate performance at 21.4%. SVM showed our method to be the best. For the next three components, KNN showed that our method was better than the other methods, but SVM indicated that both LDA and SDA were better than our algorithm. In the fifth component, KNN again showed that our method was the best. SVM showed that MRMI was better and placed our algorithm second in rank. For the last two components, both classifiers returned the result that MRMI was the best method; our method ranked second again. In this data set, our method was generally adequate in all components in comparison to the other methods. Whenever our method was ranked second, its performance was close to the first-ranked algorithm.

Table 2. Percentile average classification accuracy with KNN on the different data sets

**Letter data set**

| Dim. | Raw | PCA | LDA | SDA | MRMI | MIDA |
|---|---|---|---|---|---|---|
| 1 | 4.4 | 15.2 | 22.1 | **22.2** | 16.3 | 21.4 |
| 2 | 6.4 | 21.1 | 40.0 | 40.1 | 24.8 | **49.9** |
| 3 | 10.3 | 33.7 | 51.7 | 51.7 | 40.6 | **65.9** |
| 4 | 13.6 | 53.8 | 67.0 | 67.0 | 60.0 | **70.1** |
| 5 | 20.6 | 68.3 | 74.4 | 74.4 | 74.3 | **77.7** |
| 6 | 30.2 | 77.1 | 81.6 | 81.6 | **84.1** | 83.4 |
| 7 | 45.9 | 85.9 | 85.8 | 85.8 | **90.7** | 90.0 |

**Libras movement data set**

| Dim. | Raw | PCA | LDA | SDA | MRMI | MIDA |
|---|---|---|---|---|---|---|
| 1 | 24.4 | 23.3 | **31.4** | 26.4 | 20.3 | 26.9 |
| 2 | 46.1 | 33.1 | **50.0** | 43.6 | 26.7 | 48.1 |
| 3 | 45.3 | 50.6 | 50.6 | 56.1 | 27.8 | **66.7** |
| 4 | 46.7 | 65.0 | 55.8 | 67.8 | 28.6 | **72.5** |
| 5 | 46.1 | 71.7 | 61.9 | 76.4 | 35.3 | **78.3** |
| 6 | 46.4 | 78.6 | 62.2 | **82.8** | 35.8 | 80.3 |
| 7 | 45.6 | 81.9 | 63.6 | **85.3** | 36.1 | 82.2 |

**Wall-Following data set**

| Dim. | Raw | PCA | LDA | SDA | MRMI | MIDA |
|---|---|---|---|---|---|---|
| 1 | 50.0 | 39.3 | 49.7 | 45.2 | 40.4 | **55.3** |
| 2 | 72.2 | 54.1 | 65.9 | 64.0 | 56.5 | **74.1** |
| 3 | 81.8 | 72.5 | 75.2 | 77.1 | 68.7 | **85.2** |
| 4 | 85.0 | 80.2 | - | 81.8 | 75.3 | **89.6** |
| 5 | 85.7 | 84.3 | - | 84.7 | 80.0 | **91.3** |
| 6 | 85.2 | 87.2 | - | 85.8 | 82.7 | **91.6** |
| 7 | 84.2 | 87.3 | - | 86.8 | 84.4 | **92.1** |

**Madelon data set**

| Dim. | Raw | PCA | LDA | SDA | MRMI | MIDA |
|---|---|---|---|---|---|---|
| 1 | 49.6 | 50.2 | **53.4** | 51.0 | 51.0 | 49.5 |
| 2 | 52.0 | 54.2 | - | - | 50.1 | **54.9** |
| 3 | 50.5 | 57.2 | - | - | 50.4 | **61.9** |
| 4 | 51.3 | 64.9 | - | - | 50.6 | **74.3** |
| 5 | 49.4 | 79.5 | - | - | 48.5 | **84.2** |
| 6 | 50.7 | **88.8** | - | - | 48.0 | 87.5 |
| 7 | 51.9 | 85.9 | - | - | 47.7 | **87.6** |

**Hill-valley with noise data set**

| Dim. | Raw | PCA | LDA | SDA | MRMI | MIDA |
|---|---|---|---|---|---|---|
| 1 | 53.3 | 52.7 | **75.6** | 51.8 | 50.3 | 54.4 |
| 2 | 51.3 | 58.4 | - | 54.0 | 52.1 | **78.9** |
| 3 | 53.4 | 68.1 | - | 69.2 | 51.6 | **89.3** |
| 4 | 51.6 | 89.8 | - | 88.1 | 53.7 | **94.8** |
| 5 | 51.6 | 96.1 | - | 95.1 | 56.7 | **98.8** |
| 6 | 50.1 | 98.3 | - | - | 58.5 | **99.3** |
| 7 | 51.7 | 98.4 | - | - | 58.0 | **99.5** |

**Hill-valley without noise data set**

| Dim. | Raw | PCA | LDA | SDA | MRMI | MIDA |
|---|---|---|---|---|---|---|
| 1 | 53.1 | 49.3 | **76.1** | 52.0 | 48.9 | 50.6 |
| 2 | 54.5 | 55.9 | - | **67.2** | 49.8 | 66.8 |
| 3 | 53.3 | 75.3 | - | 75.4 | 52.8 | **88.5** |
| 4 | 54.0 | 88.7 | - | 86.2 | 55.6 | **92.5** |
| 5 | 53.9 | 95.3 | - | 94.9 | 57.8 | **97.4** |
| 6 | 53.9 | 97.4 | - | - | 58.4 | **98.4** |
| 7 | 53.2 | 98.1 | - | - | 59.5 | **99.3** |

The second data set was the Libras movement data set that contains 15 classes and 24 instances. Each class refers to a hand movement type in Libras. This data set represents the coordinates of movements with 90 features. In this data set, the KNN and SVM classifiers showed that the results obtained from our method generally were better than the results of other methods. For the first two components, the KNN classifier showed that the LDA method was better and our method ranked second. Meanwhile, the SVM classifier ranked our method first for components 3 to 5. Both classifiers showed our method to be better than the other methods. However, for the last two components, SDA performed better. For these two components, our method ranked second again.

The third data set was a wall-following robot navigation data set that consists of 24 features. Its objective is to test the hypothesis that a seemingly simple navigation task is actually a nonlinearly separable classification task. Therefore, linear classifiers, unlike non-linear classifiers, cannot be trained to perform navigations around a room without collisions. Here, the results from the KNN and SVM classifiers showed that our feature extraction method was able to obtain better discriminative information compared to the other five methods for the first seven components. Other methods could not extract the discriminative information because this data set contains features that are nonlinearly separable.

Table 3. Percentile average classification accuracy with SVM on the different data sets

**Letter data set**

| Dim. | Raw | PCA | LDA | SDA | MRMI | MIDA |
|---|---|---|---|---|---|---|
| 1 | 7.5 | 8.1 | 19.3 | 19.3 | 12.1 | **20.9** |
| 2 | 8.8 | 19.0 | **43.5** | **43.5** | 26.6 | 37.9 |
| 3 | 12.8 | 36.4 | **56.0** | **56.0** | 44.8 | 53.5 |
| 4 | 15.8 | 59.1 | **70.4** | **70.4** | 61.4 | 62.7 |
| 5 | 23.0 | 72.9 | 76.7 | 76.7 | **77.30** | 76.8 |
| 6 | 34.3 | 79.8 | 83.7 | 83.7 | **87.0** | 85.6 |
| 7 | 51.6 | 84.2 | 87.5 | 87.5 | **92.6** | 91.8 |

**Libras movement data set**

| Dim. | Raw | PCA | LDA | SDA | MRMI | MIDA |
|---|---|---|---|---|---|---|
| 1 | 15.3 | 9.4 | 19.2 | 16.4 | 7.5 | **22.8** |
| 2 | 32.2 | 28.1 | 40.3 | 33.9 | 15.8 | **43.6** |
| 3 | 33.9 | 45.8 | 48.9 | 44.4 | 21.4 | **60.0** |
| 4 | 35.3 | 61.9 | 51.4 | 60.0 | 26.1 | **70.0** |
| 5 | 35.3 | 67.8 | 54.4 | 69.2 | 31.1 | **71.9** |
| 6 | 35.6 | 71.4 | 55.8 | **73.9** | 34.2 | 73.3 |
| 7 | 36.7 | 72.5 | 57.5 | **77.8** | 36.4 | 74.2 |

**Wall-Following data set**

| Dim. | Raw | PCA | LDA | SDA | MRMI | MIDA |
|---|---|---|---|---|---|---|
| 1 | 55.7 | 44.5 | 62.3 | 55.8 | 48.3 | **66.9** |
| 2 | 69.1 | 52.8 | 68.2 | 64.9 | 58.1 | **70.7** |
| 3 | 79.7 | 67.6 | 73.3 | 75.5 | 66.5 | **80.0** |
| 4 | 83.3 | 76.3 | - | 78.0 | 75.1 | **85.1** |
| 5 | 84.5 | 83.1 | - | 81.8 | 80.2 | **89.1** |
| 6 | 82.8 | 86.7 | - | 83.7 | 82.0 | **90.3** |
| 7 | 80.9 | 86.9 | - | 86.5 | 82.5 | **91.5** |

**Madelon data set**

| Dim. | Raw | PCA | LDA | SDA | MRMI | MIDA |
|---|---|---|---|---|---|---|
| 1 | 46.5 | 46.8 | 53.7 | **55.4** | 46.7 | 55.2 |
| 2 | 46.3 | 55.5 | - | - | 48.8 | **60.1** |
| 3 | 48.1 | 63.8 | - | - | 49.1 | **66.9** |
| 4 | 47.7 | 71.2 | - | - | 48.7 | **78.7** |
| 5 | 51.7 | 83.6 | - | - | 49.9 | **88.2** |
| 6 | 50.9 | **91.4** | - | - | 50.5 | 89.2 |
| 7 | 52.4 | 89.4 | - | - | 50.5 | **90.0** |

**Hill-valley with noise data set**

| Dim. | Raw | PCA | LDA | SDA | MRMI | MIDA |
|---|---|---|---|---|---|---|
| 1 | 47.6 | 45.7 | **68.9** | 53.3 | 49.3 | 54.0 |
| 2 | 47.4 | 49.7 | - | **54.3** | 47.7 | 53.4 |
| 3 | 47.6 | 50.8 | - | 56.0 | 49.8 | **56.9** |
| 4 | 47.5 | 53.6 | - | 59.1 | 49.2 | **60.2** |
| 5 | 47.3 | 55.2 | - | 57.9 | 50.4 | **63.4** |
| 6 | 47.4 | 55.9 | - | - | 49.0 | **64.3** |
| 7 | 47.0 | 57.4 | - | - | 49.3 | **64.9** |

**Hill-valley without noise data set**

| Dim. | Raw | PCA | LDA | SDA | MRMI | MIDA |
|---|---|---|---|---|---|---|
| 1 | 49.1 | 50.9 | **69.3** | 51.6 | 49.3 | 51.8 |
| 2 | 49.1 | 53.6 | - | **56.6** | 49.9 | 56.1 |
| 3 | 49.1 | 55.2 | - | 57.7 | 49.6 | **63.4** |
| 4 | 49.4 | 58.4 | - | 57.8 | 49.5 | **65.8** |
| 5 | 49.8 | 62.0 | - | 61.0 | 49.2 | **65.4** |
| 6 | 50.1 | 62.5 | - | - | 49.9 | **66.0** |
| 7 | 50.7 | 64.3 | - | - | 49.3 | **66.6** |

The fourth data set used in our paper was Madelon, which is an artificial data set containing data points that are divided into 32 clusters. Each cluster is located at the vertices of a five dimensional hypercube and randomly labeled +1 or -1. This data set is constructed of 500 attributes. Of these attributes, 5 are relative features, 15 are redundant features consisting of linear combinations of those relative features, and the remaining 480 features are irrelevant. In this case, our method was better based on the KNN and SVM classifier results. The detailed results are as follows: for the first component, KNN indicated that the LDA method was better. The SVM classifier ranked

SDA first and our method second. For the remaining components, with the exception of the sixth component, both the KNN and SVM classifiers indicated that our algorithm was the best. For the sixth component, both classifiers showed that the PCA method was better, and our method ranked second again.

The fifth and sixth data sets are the Hill-valley data set. This data set is made up of 100 points on a two-dimensional graph. Its objective is to distinguish hills from valleys, and it consists of two data sets. One data set is noisy, and the other data set is noiseless. We used both sets in our experiment. In the fifth case, using the noisy subdata, our method was generally better according to the KNN classifier results. For the first component, LDA performed better than the others, and ours ranked second. For the rest of components, our method had better performance. However, according to the results obtained from the SVM classifier for the first component, LDA was the leading method, and our method ranked second. For the second component, SDA was better and we ranked second. For the rest of the components, we ranked first. For this data set, according to the results that we obtained from the KNN classifier, our method was the best. The SVM classifier showed that LDA was better, but if we compare all results together, our method was generally the leading method. In the sixth case, using the second subdata (the set without noise), our method again ranked first as the best method for extracting discriminative information. For the first component, according to KNN classifier, the result of LDA method was the best. However, for the second component, SDA ranked first place with 67.2%. We ranked in second place with 66.8%. For the remaining components, our method ranked first. Meanwhile, for the first component, according to the SVM classifier, the leading method was LDA. For the second component, the leading method was SDA. For both components, our method ranked second. However, our method ranked first for the remaining components. Again, as we observed in the previous data set, KNN ranked our method highest. SVM ranked LDA above our method, but in aggregate, our method was better.

## 5. CONCLUSIONS

Feature extraction plays an important role in classification systems. In this paper, a novel DA method for feature extraction based on MI was proposed. This method is called MIDA. The goal of MIDA is to create new features by transforming the original features such that the transformation simultaneously maximizes the MI between the transformed features and the class labels and minimizes redundancy. In contrast to other DA algorithms that are based on second-order statistics, the proposed method is based on information theory that is able to compare the nonlinear relationships between random variables (i.e., between a vector of features and the class label). The proposed method was evaluated using six data sets from UCI databases. The experiments have shown that the MIDA method performs comparably to existing methods. MIDA's performance was frequently better than other methods. For experiments when it was not ranked the best, MIDA ranked close to the best. MIDA is effective for dimensionality reduction.

## Authors

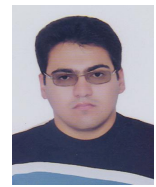

Ali Shadvar was born in Bandaanzali, Iran in 1981. He received the B.Sc. degree in biomedical engineering from Sahand University of Technology, Tabriz, Iran, in 2005, and the M.Sc. degree in biomedical engineering from Iran University of Science and Technology (IUST), Tehran, in 2010. He is currently an Assistant Professor at the University Islamic Azad University, Langroud, Iran. His research involves data mining, feature extraction and selection, Discriminant analysis, Dimension reduction, pattern recognition, machine learning and rough set.